\documentclass{article}

     \PassOptionsToPackage{numbers}{natbib}
     \usepackage[preprint]{neurips_2021}

\usepackage[utf8]{inputenc} % allow utf-8 input
\usepackage[T1]{fontenc}    % use 8-bit T1 fonts
\usepackage{hyperref}       % hyperlinks
\usepackage{url}            % simple URL typesetting
\usepackage{booktabs}       % professional-quality tables
\usepackage{amsfonts}       % blackboard math symbols
\usepackage{nicefrac}       % compact symbols for 1/2, etc.
\usepackage{microtype}      % microtypography
\usepackage{xcolor}         % colors

\usepackage{multirow}
\usepackage{subfigure}
\usepackage{graphicx}
\usepackage{tabularx}
\usepackage{amsmath}

\title{When Liebig's Barrel Meets Facial Landmark Detection: A Practical Model}

\author{Haibo Jin\thanks{Equal contribution}, \hspace{1mm} Jinpeng Li\footnotemark[1], \hspace{1mm} Shengcai Liao\thanks{Corresponding author} \hspace{1mm} \& \hspace{1mm} Ling Shao \\
Inception Institute of Artificial Intelligence (IIAI), Abu Dhabi, UAE\\
\texttt{haibo.nick.jin@gmail.com, ljpadam@gmail.com} \\
\texttt{ \{scliao,ling.shao\}@ieee.org} \\
}

\begin{document}

\maketitle

\begin{abstract}
In recent years, significant progress has been made in the research of facial landmark detection. However, few prior works have thoroughly discussed about models for practical applications. Instead, they often focus on improving a couple of issues at a time while ignoring the others. To bridge this gap, we aim to explore a practical model that is accurate, robust, efficient, generalizable, and end-to-end trainable at the same time. To this end, we first propose a baseline model equipped with one transformer decoder as detection head. In order to achieve a better accuracy, we further propose two lightweight modules, namely dynamic query initialization (DQInit) and query-aware memory (QAMem). Specifically, DQInit dynamically initializes the queries of decoder from the inputs, enabling the model to achieve as good accuracy as the ones with multiple decoder layers. QAMem is designed to enhance the discriminative ability of queries on low-resolution feature maps by assigning separate memory values to each query rather than a shared one. With the help of QAMem, our model removes the dependence on high-resolution feature maps and is still able to obtain superior accuracy. Extensive experiments and analysis on three popular benchmarks show the effectiveness and practical advantages of the proposed model. Notably, our model achieves new state of the art on WFLW as well as competitive results on 300W and COFW, while still running at 50+ FPS. 
\end{abstract}

\section{Introduction}
\label{sec:1}

Facial landmark detection is an essential preceding task of several computer vision problems, such as face recognition~\citep{TYR14, LWY17, LJL13}, face tracking~\citep{KMT17}, and face editing~\citep{TZS16}. Thanks to convolutional neural networks (CNNs), we have witnessed significant progresses in facial landmark detection recently. Despite its fast development, the existing models often focus on a couple of issues at a time while ignoring the others, which seems to be inappropriate for practical applications. According to Liebig's law, the capacity of a barrel is limited by its shortest stave. That is to say, the lengths of all the staves should be considered comprehensively when designing a barrel. Similarly, such a law also applies to facial landmark detectors. 

We propose to assess the overall performance of a facial landmark detector from five aspects, namely (1) \textbf{accuracy}; (2) \textbf{structural robustness}; (3) \textbf{efficiency}; (4) \textbf{generalization}; and (5) \textbf{ease of training}. Accuracy is indubitably the main concern of a facial landmark detection model as it measures how good a model's predictions are on average. However, a model's predictions may also be considered unsatisfactory if some of the predicted landmarks are greatly deviated from the ground-truths, even if the accuracy is good on average. Therefore, structural robustness is a necessary metric for a practical model. Besides, face-related applications (e.g., identity verification and virtual makeup) are often real-time systems, thus requiring high computational efficiency, which also should be a factor to consider. Furthermore, generalization capability is essential for practical models because they need to run in the wild. Last but not least, ease of training should play an important role in practice, but it usually draws little attention from researchers. We consider a model to be easy to train if it can be trained end-to-end with respect to ground-truths and contains as few hyperparameters as possible.

In this work, we aim to explore a facial landmark detector that comprehensively considers the above five aspects. To this end, we first propose a baseline model based on transformers due to its superior capability on dynamic feature extraction and global attentions. We adopt the transformers from \citep{CMS20} by setting the number of queries to be $N$, where $N$ is the number of landmarks. At the prediction layer, each query predicts its corresponding two-dimensional coordinates. We also remove the encoder based on the experimental results in Section~\ref{sec:4.3}. Besides, there are experiments showing that the accuracy of our baseline model can be improved by stacking more decoder layers or increasing the resolutions of feature maps. However, the computational cost also considerably increases as a result. Moreover, it brings extra hyperparameters, i.e., the number of decoder layers and the stride of backbone. Therefore, we decide to equip the baseline model with only one decoder layer and relatively low-resolution feature maps (e.g., stride 32). With the above modifications, we obtain a baseline model that is potentially efficient, robust, generalizable, and end-to-end trainable.

To further improve the accuracy without adding too much computational burden, we propose two lightweight modules, namely dynamic query initialization (DQInit) and query-aware memory (QAMem). The two modules are designed to make up the accuracy loss from fewer decoder layers and lower feature map resolutions respectively. To be more specific, we argue that the queries can be initialized with a coarse estimate from inputs rather than zeros as in \citep{CMS20}. DQInit takes the memory as input and outputs $N$ query embeddings through a global average pooling layer and a fully connected layer, which is computationally efficient. In terms of low-resolution feature maps, we have also observed a deficiency of the transformer in baseline model, i.e., the queries are not able to extract different values within the same grid of the memory, given the same attention weights. This is because all the queries share the same memory, and it does not cause a problem when it uses high-resolution maps as it can access more refined spatial features to distinguish different queries. However, the shared memory reduces the discriminative ability of the queries on low-resolution maps. In response to this problem, we propose QAMem, where each query has its own memory values. As we show in Section~\ref{sec:3.3}, QAMem can be efficiently implemented with a 1x1 group convolutional layer over query embeddings. We refer to the baseline model equipped with DQInit and QAMem as BarrelNet. The effectiveness of BarrelNet has been verified through extensive experiments on three popular benchmarks. 

Our contributions can be summrized as follows.

\begin{enumerate}
\item We propose to comprehensively assess a facial landmark detector from five aspects, which is more appropriate for practical applications. Accordingly, we propose a transformer-based baseline model that is potentially efficient, robust, generalizable, and end-to-end trainable.
\item A lightweight DQInit module is designed to boost the accuracy of baseline model so that the number of decoder layers can be reduced to one while the accuracy almost does not drop.
\item A novel QAMem module is proposed to further improve the accuracy so that high-resolution feature maps are no longer necessary for obtaining superior accuracy. 
\item The proposed BarrelNet achieves new state of the art on WFLW as well as competitive results on 300W and COFW. The effectiveness and practical advantages of BarrelNet have been thoroughly analyzed through experiments.
\end{enumerate}

\section{Related Work}
\label{sec:2}

%In this section, we review relevant works on facial landmark detection and visual transformers. 

\subsection{Facial landmark detection}

According to the type of regression, facial landmark detectors can be categorized as coordinate regression-based and heatmap regression-based models. 

\begin{figure}
\centering
  \includegraphics[width=0.9\linewidth]{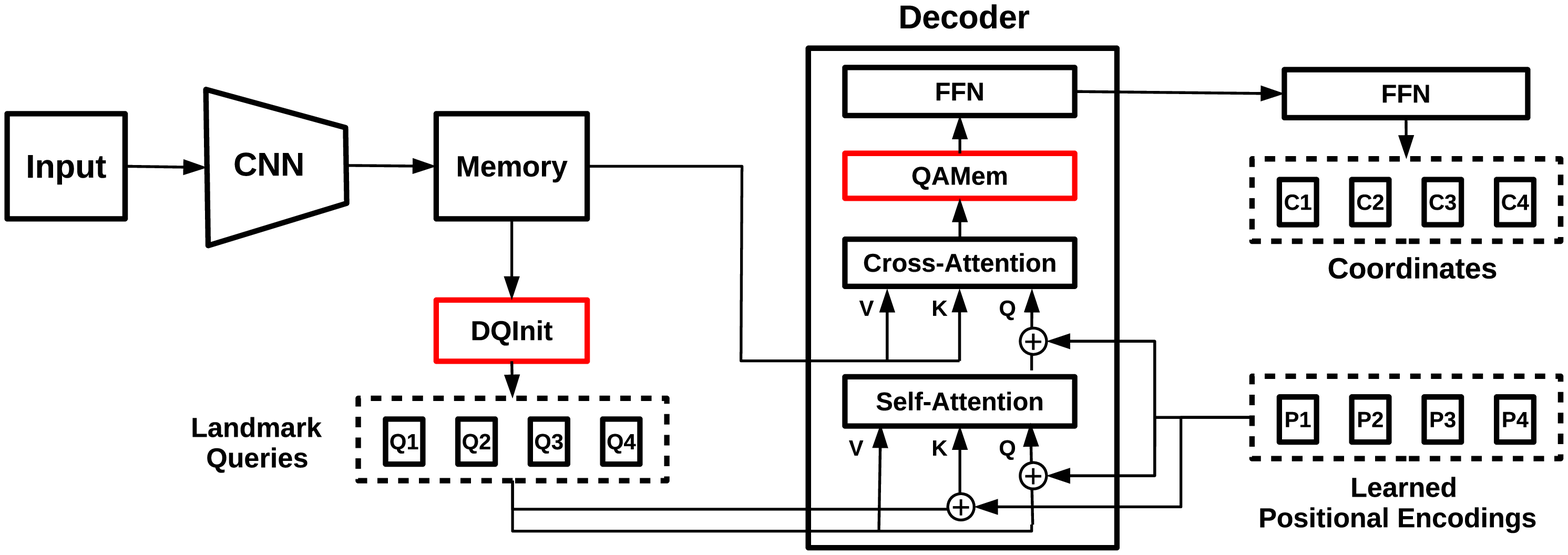}
\vspace{-1mm}
\caption{The architecture of BarrelNet, which mainly consists of a CNN as the backbone and a decoder layer as the detection head. The two modules in red, DQInit and QAMem, are the key components of BarrelNet for a better accuracy and efficiency. Here we use 4 landmarks as an example.}
\label{fig:framework}      
\end{figure}

Coordinate regression-based models~\citep{WQY18,ZSZ19,VBV18,LZH19} directly regress the coordinates of landmarks, thus being end-to-end trainable. Fully connected layers are usually used for the implementation of coordinate regression, which are not accurate enough due to the fixed weights. To localize landmarks more precisely and robustly, coordinate regressions can be used in a coarse-to-fine manner through cascaded structures~\citep{SWT13,ZLL15,TSN16,LSX17,FKA18}. DAG~\citep{LLZ20} is a recent work that applied graph convolutional networks to landmark detection, enabling the model to dynamically leverage global and local features with graph signals. Despite the superior performance, DAG is computationally inefficient as it relies on high-resolution feature maps and multi-stage regressions. 

Heatmap regression-based models use heatmaps as a proxy of ground-truths, where each heatmap represents the probability of a landmark position. These models cannot be trained in an end-to-end manner, and they usually introduce label-related hyperparameters, which is not training-friendly. Typically, heatmap regression is based on high-resolution feature maps for precise localization. Stacked hourglass network~\citep{NYD16,YLZ17,LZH19,CSJ19,DoY19,ZZY19,WBF19,CBG20} and U-Net~\citep{RFB15,TPG18,ZZY19,DBC19,KMM20} are two widely used architectures among such models. HRNet~\citep{WSC19} is a recently proposed architecture that obtains high-resolution maps by maintaining feature maps with different resolutions in parallel and constantly exchanging information between them. More recently, PIPNet~\citep{JLS20} was proposed as an efficient model since it only utilizes low-resolution feature maps to achieve competitive accuracy. PIPNet is also equipped with a neighbor regression module to yield structural robustness on its predictions. 

In addition to accuracy, efficiency, and robustness, generalization is also a key issue for real applications. \citet{DYO18} and \citet{QSW19} utilized style transfer techniques to deal with style changes. \citet{JLS20} improved generalization capability of PIPNet by designing a generalizable semi-supervised learning method. DAG~\citep{LLZ20} performed well on cross-domain evaluation due to the dynamic mechanism from graph message passing. 

The existing works have discussed every important aspect of facial landmark detection, but few of them have explored a practical model that meets all of them. Instead, they often focus on a couple of issues while ignoring the others. Take the recent works for example, DAG is accurate, structural robust, and generalizable, but it relies on high-resolution feature maps, which could be an obstacle to practical use; PIPNet is fast, accurate, and robust, but it is not end-to-end trainable. Besides, there is a performance gap between PIPNet and state-of-the-art methods. Different from prior works, we aim to explore a practical model that meets as many aspects as possible while being accurate. 

\subsection{Transformers for vision tasks}

Transformers were first proposed and applied in natural language processing~\citep{VSP17}, then adopted to computer vision tasks. ViT~\citep{DBK21} introduces transformers to replace CNNs as a backbone, obtaining superior performance on multiple image classifiction benchmarks. DETR~\citep{CMS20} adopted transformers for object detection and panoptic segmentation, obtaining an end-to-end model that achieves competitive results on the two tasks. Deformable DETR~\citep{ZSL21} is a follow-up work of DETR, which significantly reduces the training time and improves the performance on small objects. Please refer to \citep{KNH21} for a more detailed survey on visual transformers. In this work, we adopt the transformer from DETR and apply it to facial landmark detection with multiple task-specific modifications.  

\begin{figure}
\centering
    \subfigure[\label{fig:QAMem_a}]{
    \includegraphics[width=0.22\linewidth]{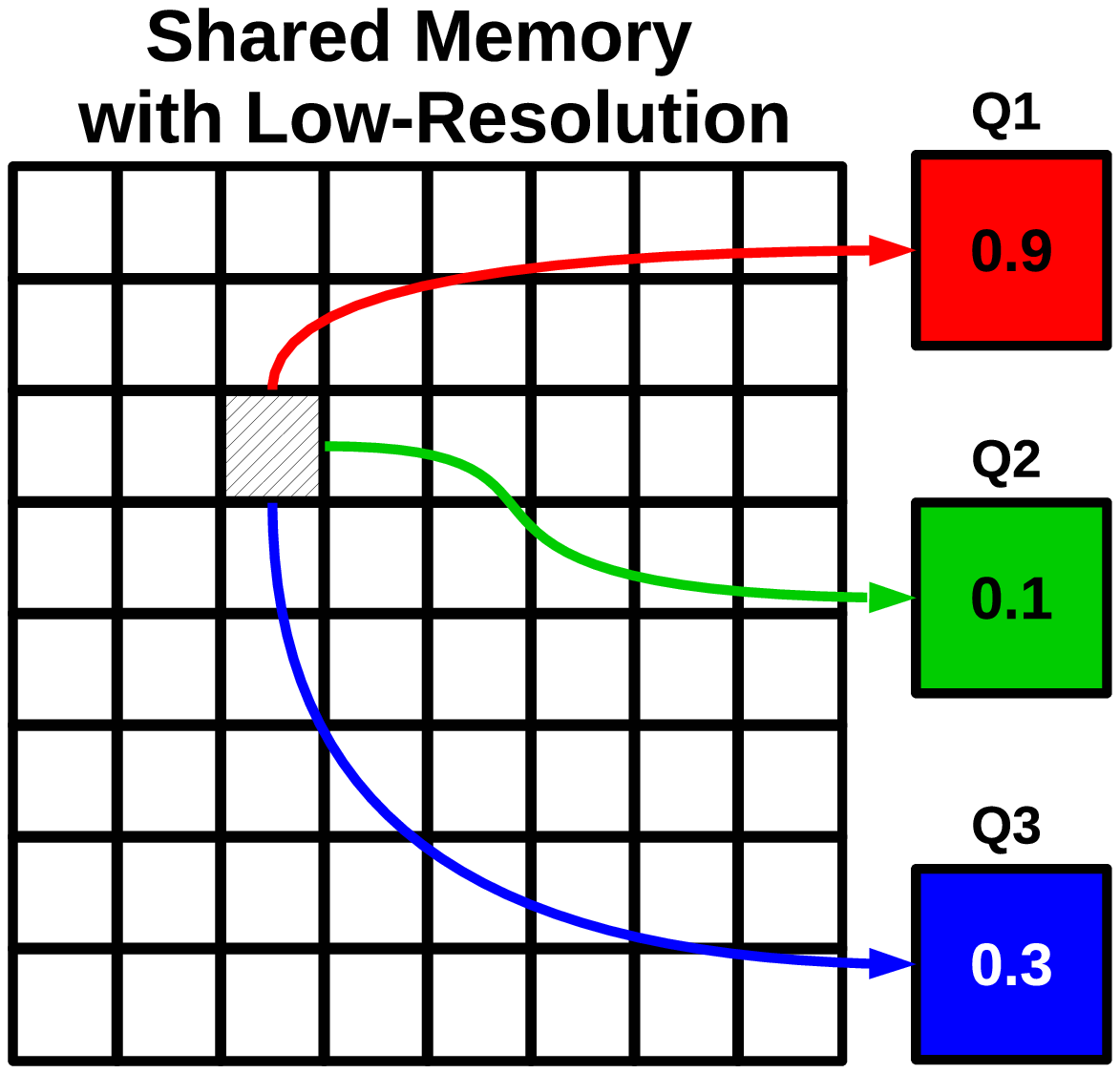}
    } 
    \subfigure[\label{fig:QAMem_b}]{
    \includegraphics[width=0.22\linewidth]{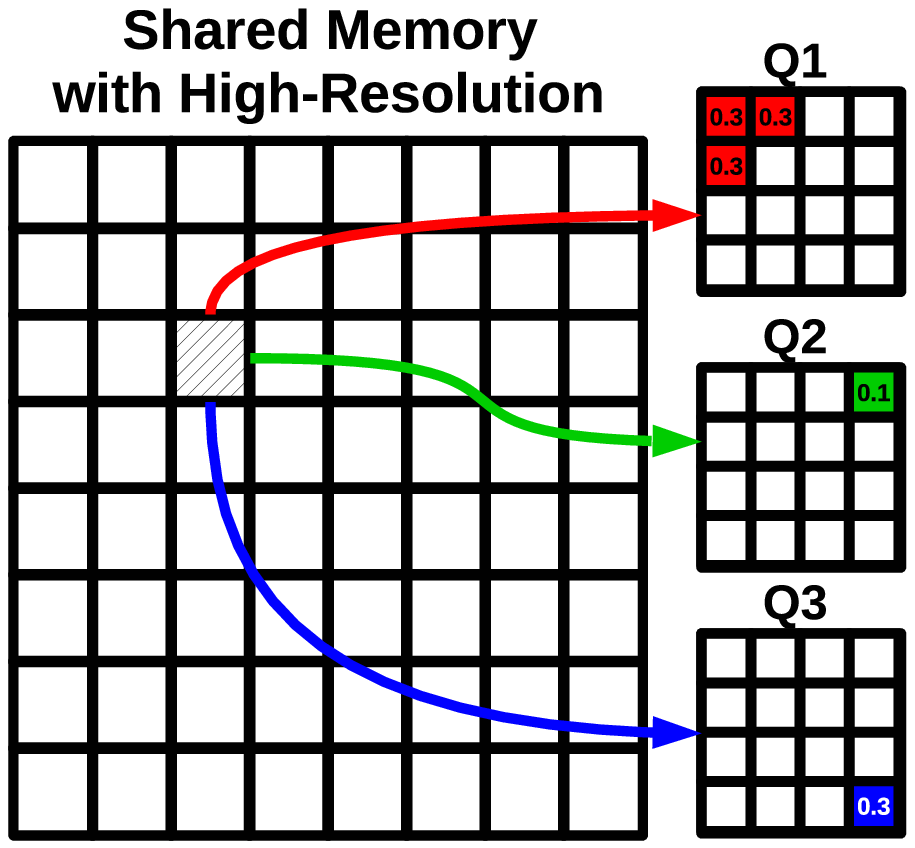}
    }     
    \subfigure[\label{fig:QAMem_c}]{
    \includegraphics[width=0.51\linewidth]{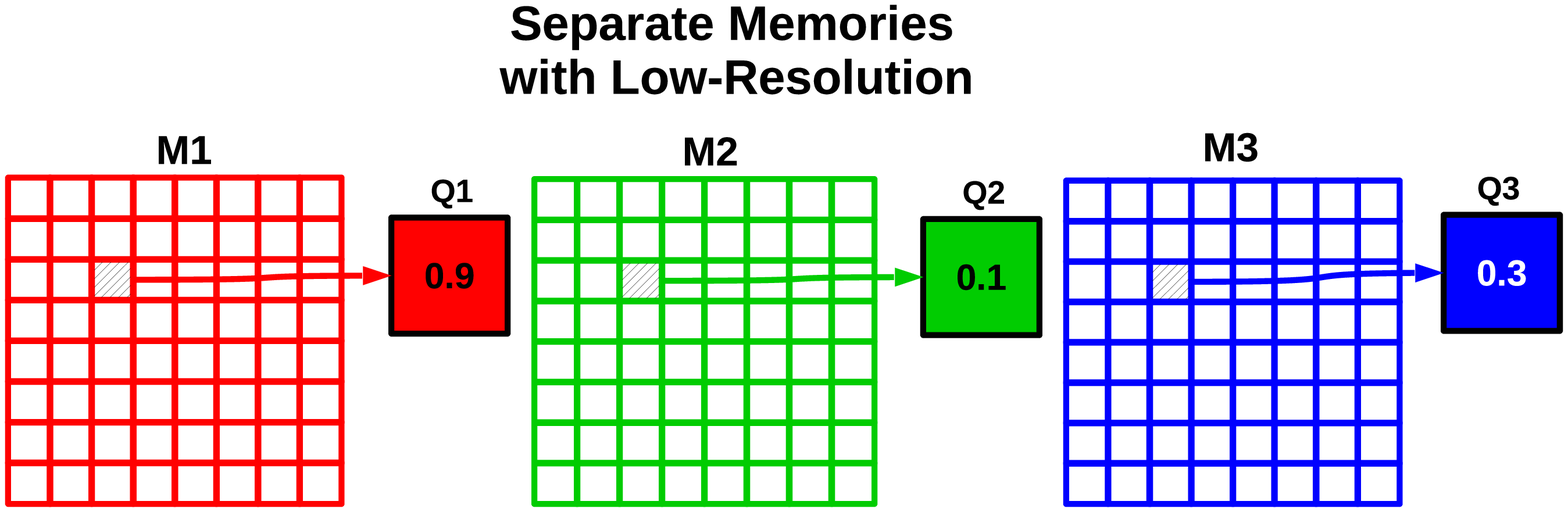}
    }
    \vspace{-3mm}
    \caption{Different memory mechanisms. (a) A shared memory with low-resolution maps. (b) A shared memory with high-resolution maps. (c) Separate memories with low-resolution feature maps.}
    \label{fig:QAMem}
\end{figure}

\section{Method}
\label{sec:3}

In this section, we first introduce the baseline model in \ref{sec:3.1}, then the two modules DQInit and QAMem in \ref{sec:3.2} and \ref{sec:3.3}, respectively. 

\subsection{Transformers for facial landmark detection}
\label{sec:3.1}

The transformer in DETR~\citep{CMS20} first passes the feature maps from backbone to a stacked encoder to obtain a memory, then decodes the memory with a stacked decoder and learned positional encodings. There are $N$ zero-initialized object queries as the inputs of decoder, then tranformed to $N$ embeddings as the outputs, where $N$ is usually a much larger number (e.g., 300) than the number of objects in an image. Finally, each embedding predicts a class label and bounding box coordinates.

To adapt the transformer to a baseline model of facial landmark detection, we make three modifications. Firstly, we set $N$ to be the number of landmarks, and  the output embeddings only predict landmark coordinates. Secondly, we remove the encoder as the experiments in Section~\ref{sec:4.3} show that encoder is not as helpful as decoder in our task. Lastly, we reduce the number of decoder layers to one to save computational costs. The framework of the baseline model can be seen in Figure~\ref{fig:framework} excluding the red colored parts, which are newly designed modules on the basis of baseline.

The baseline model is a good starting point because it is end-to-end trainable with minimum hyperparameters and has the potential to perform well in terms of structural robustness and generalization. To be specific, the self-attention layers in the decoders give each query embedding access to other queries, enabling the model to make robust predictions through global shape constriants. For generalization, since transformers contain fewer inductive biases than CNNs~\citep{DBK21}, they have more freedom to learn such biases from data as long as they are well trained. Consequently, transformer-based models may benefit more from large-scale training such that it generalizes better on unseen data.

\subsection{Dynamic query initialization}
\label{sec:3.2}

Intuitively, the learned positional encodings can be seen as a prior that the model learned from data since they are fixed after training. In contrast, the landmark queries are data-dependent because their values need to be extracted from the memory during inference. Although it works well when the queries are zero-initialized as in \citep{CMS20}, we argue that the performance can be further improved if the landmark queries are initialized with more meaningful values, especially when there is only one decoder layer. Accordingly, we propose a simple and lightweight module named DQInit, which only consists of a global average pooling layer and a fully connected layer. Suppose the memory is denoted as $M$ with size $d \times h \times w$ and each query embedding is of size $d$, the dynamic initialization of queries can be computed as follows.

\begin{equation}
Q_{\text{init}} = \text{FC}(\text{GlobalAvgPool}(M)),
\end{equation}

where $Q_{\text{init}}$ is of size $N \times d$, with $N$ being the number of queries.

\subsection{Query-aware memory}
\label{sec:3.3}

The experiments in Section~\ref{sec:4.3} show that higher-resolution feature maps (i.e., smaller strides) bring better accuracy, but also introduce heavy computations, which is not friendly to practical applications. In contrast, low-resolution maps are efficient but the accuracy is degraded. To address this trade-off problem, we propose a novel and lightweight module named QAMem based on an observation. To be specific, we observe that the discriminative ability of the queries is limited within a specific grid of the memory. Suppose there are three queries $Q_1$, $Q_2$, and $Q_3$, and the embedding value of the grid is denoted as $V_g$. As can be seen in Figure~\ref{fig:QAMem_a}, the resolution of the memory is $8 \times 8$ and the attention weights of the three queries are $0.9$, $0.1$, and $0.3$ respectively. Then we can get the extracted value from the grid for each query, i.e., $0.9V_g$, $0.1V_g$, and $0.3V_g$ respectively. As can be seen, the only difference of the three values is the scale. On the other hand, if the memory is of higher resolutions, then the queries will act differently. In Figure~\ref{fig:QAMem_b}, the memory is $32 \times 32$, so the previous grid becomes a $4 \times 4$ sub-map. Due to the increased resolutions, each query now is able to extract a different value from the grid. Can we also obtain such discriminative abilities on low-resolution maps? Yes, if the memory act differently to different queries. To achieve this, we compute $N$ new memories for $N$ queries from the original memory through $N$ convolutional layers, where each query has a corresponding convolutional layer. Figure~\ref{fig:QAMem_c} shows the mechanism that each query has a separate memory, which enables different queries to extract different values from the same grid on low-resolution maps. Note that the separate memories are only for value extractions, while the key of the queries is still shared. 

Nevertheless, simply implementing the above method can be computationally heavy since there are usually tens of landmarks thus as many new memories to generate. To address this, we propose an equivalent implementation that is much more efficient. Specifically, suppose $A$ is the attention weights with size $N \times S$, $M$ is the original memory with size $S \times d$, $T^i$ is the corresponding transform (i.e., conv layer) of $Q_i$ with size $d \times d$, where $S$ denotes the number of feature map grids (i.e., $hw$). Then the extracted query $Q_i$ from the above method can be computed as follows:

\begin{equation}
Q_i = A_{N \times S} \cdot M^i_{S \times d} = A_{N \times S} \cdot (M_{S \times d} \cdot T^i_{d \times d}).
\end{equation} 

%\begin{equation}
%Q_i = \underset{(N \times S)}{A} \cdot \underset{(S \times d)}{M^i} = \underset{(N \times S)}{A} \cdot (\underset{(S \times d)}{M} \cdot \underset{(d \times d)}{T^i})
%\end{equation}

With the simplified implementation, query $Q_i$ can be computed as: 

\begin{equation}
Q_i = A_{N \times S} \cdot (M_{S \times d} \cdot T^i_{d \times d}) = (A_{N \times S} \cdot M_{S \times d}) \cdot T^i_{d \times d}.
\end{equation} 

After simplification, the transform is only computed over the extracted query embedding rather than memory, which significantly reduces redundant computations and memories. In practice, the QAMem layer can be simply implemented as a convolutional layer with $1 \times 1$ kernel, stride $1$, and $N$ groups.  

\section{Experiments}
\label{sec:4}

In this section, we first illustrate the settings for experiments in Section~\ref{sec:4.1}, then present the results of our model with a comparison to state-of-the-art methods in Section~\ref{sec:4.2}. Finally, we conduct detailed analysis of the proposed model in Section~\ref{sec:4.3}. 

\subsection{Experimental settings}
\label{sec:4.1}

\paragraph{Datasets.} 
Three popular benchmarks are used in our experiments, namely 300W, COFW, and WFLW. \textbf{300W}~\citep{STZ13} includes 3,148 training images and 689 test images, where each image is annotated with 68 landmarks. The test set can be further divided into two subsets, common set and challenging set, containing 554 and 135 images respectively. \textbf{COFW}~\citep{BPD13} is a dataset with heavy occlusions, containing 1,345 training images and 507 test images. COFW was originally annotated with 29 landmarks, then re-annotated~\citep{GhF14} with 68 landmarks (only test set) for an easy cross-dataset evaluation from 300W. For COFW, we report results on both annotations, and we denote the 68-landmark version as COFW-68. \textbf{WFLW}~\citep{WQY18} was collected from WIDER Face~\citep{YLL16}, annotated with 98 landmarks. It contains 7,500 and 2,500 images for training and testing, respectively. Due to the diverse scenarios, WFLW is an appropriate dataset for the evaluation of practical models. 

\begin{table}
\centering
\caption{Comparison with state-of-the-art methods on 300W, COFW, and WFLW. The results are in NME (\%), using inter-ocular distance for normalization.  \textcolor{red}{Red} indicates best, and \textcolor{blue}{blue} for second best.}
\newcolumntype{C}{>{\centering\arraybackslash}X}%
\begin{tabularx}{\textwidth}{lClCCCCC}
\toprule
\multirow{2}{*}{Method} & \multirow{2}{*}{Year} & \multirow{2}{*}{Backbone} & \multicolumn{3}{c}{300W} & COFW & WFLW\\ 
\cmidrule(r){4-6}
 & & & Full & Com. & Cha. & Full & Full\\
\midrule
%RCN~\citep{HYV16}           & 2016 & -  & 5.41   & 4.67   & 8.44   & -    & - \\ 
%DAC-CSR~\citep{FKC17}        & 2017 & -  & -      & -      & -      & 6.03 & -    \\ 
%TSR~\citep{LSX17}                         & 2017 & -  & 4.99   & 4.36    & 7.56    & -    & - \\
LAB~\citep{WQY18}  & 2018 & ResNet-18  & 3.49   & 2.98   & 5.19   & 5.58 & 5.27 \\ 
Wing~\citep{FKA18}           & 2018 & ResNet-50 & -   & -   & -   & 5.07    & 4.99 \\ 
SAN~\citep{DYO18}           & 2018 & ITN-CPM  & 3.98   & 3.34   & 6.60   & -  & - \\
RCN+~\citep{HMT18}           & 2018 & -  & 4.90   & 4.20   & 7.78   & -    & - \\ 
HG+SA+GHCU~\citep{LZH19}     & 2019 & Hourglass  & -      & -      & -      & -    & - \\ 
TS$^3$~\citep{DoY19} & 2019 & Hourglass+CPM & 3.78 & - & -   & -  & -    \\ 
LaplaceKL~\citep{RLZ19}      & 2019 & -  & 4.01   & 3.28   & 7.01   & -    & -  \\ 
HG-HSLE~\citep{ZZY19}        & 2019 & Hourglass  & 3.28   & 2.85   & 5.03   & -    & -    \\ 
ODN~\citep{ZSZ19}            & 2019 & ResNet-18  & 4.17   & 3.56   & 6.67   & -  & - \\ 
AVS w/ SAN~\citep{QSW19}   & 2019 & ITN-CPM  & 3.86   & 3.21   & 6.49   & -    & 4.39    \\
HRNet~\citep{WSC19}          & 2019 & HRNet-W18  & 3.32   & 2.87   & 5.15   & 3.45 & 4.60 \\ 
AWing~\citep{WBF19}       & 2019 & Hourglass & \textcolor{blue}{\textbf{3.07}} & 2.72 & \textcolor{red}{\textbf{4.52}} & - & 4.36 \\
DeCaFA~\citep{DBC19} & 2019 & Cascaded U-net & 3.69 & - & - & - & 5.01 \\ 
ADA~\citep{CBG20} & 2020 & Hourglass & 3.50 & \textcolor{red}{\textbf{2.41}} & 5.68 & - & - \\ 
LUVLi~\citep{KMM20} & 2020 & DU-Net & 3.23 & 2.76 & 5.16 & - & 4.37 \\ 
DAG~\citep{LLZ20} & 2020 & HRNet-W18 & \textcolor{red}{\textbf{3.04}} & \textcolor{blue}{\textbf{2.62}} & 4.77 & - & \textcolor{blue}{\textbf{4.21}} \\ 
PIPNet~\citep{JLS20} & 2020    & ResNet-18  & 3.36   & 2.91   & 5.18   & 3.31 & 4.57 \\ 
PIPNet & 2020    & ResNet-50  & 3.24   & 2.80   & 5.03   & 3.18 & 4.48 \\
PIPNet & 2020    & ResNet-101  & 3.19   & 2.78   & 4.89   & \textcolor{red}{\textbf{3.08}} & 4.31 \\ 
\midrule
Baseline (ours) & -    & ResNet-18  & 3.34   & 2.95   & 4.93   & 3.34 & 4.47 \\ 
Baseline (ours) & -    & ResNet-50  & 3.22   & 2.83   & 4.79   & 3.22 & 4.31 \\
Baseline (ours) & -    & ResNet-101  & 3.18   & 2.81   & 4.68   & 3.19 & 4.25 \\ 
\midrule
BarrelNet (ours) & -    & ResNet-18  & 3.22   & 2.83   & 4.84   & 3.24 & 4.42 \\ 
BarrelNet (ours) & -    & ResNet-50  & 3.14   & 2.75   & 4.71   & 3.13 & 4.24 \\
BarrelNet (ours) & -    & ResNet-101  & 3.09   & 2.73   & \textcolor{blue}{\textbf{4.60}}   & \textcolor{blue}{\textbf{3.10}} & \textcolor{red}{\textbf{4.20}} \\
\bottomrule
\end{tabularx}
\label{tab:results}
\end{table}

\paragraph{Implementation details.}
For all the datasets, we first crop the faces using the provided bounding boxes, then resize them to $256 \times 256$. For training, L1 loss is used as the loss function. Data augmentations include translation, horizontal flipping, rotation, occlusion, and blurring. We use ImageNet pretrained ResNets as our backbone by default. HRNet was also explored (see Section~\ref{sec:4.3}). BarrelNet with ResNet-50 is denoted as BarrelNet-50, and it applies to others similarly. There are 360 training epochs in total, where the learning rate is 0.0001 initially then decayed by 10 at 240th epoch. The learning rate of backbone is further multiplied by 0.1. Adam~\citep{KiB15} is used as the optimizer and the batch size is set to 16. For transformer, the hidden dimension $d$ is 256. Our code was implemented with PyTorch 1.6. For devices, we used Intel Xeon Gold 5120 @2.20GHz as CPU and NVIDIA RTX 2080Ti as GPU for both training and testing. The reported results were averaged over three runnings.

\paragraph{Evaluation metric.}
Normalized mean error (NME) is a widely used metric for landmark detection models, and it can be calculated as:

\begin{equation}
\text{NME}(y,\hat{y}) = \frac{1}{N} \sum_{i=1}^N \frac{\left\| y_i - \hat{y_i} \right\|_2}{D},
\end{equation}

where $y_i$ and $\hat{y_i}$ are ground-truth and prediction of $i$-th landmark respectively, $N$ is the number of landmarks, and $D$ is the normalization distance. In this work, we use inter-ocular as the normalization distance for all the datasets. 

\begin{figure}
\centering
    \subfigure[Occlusion\label{fig:robust_occ}]{
    \includegraphics[width=0.23\linewidth]{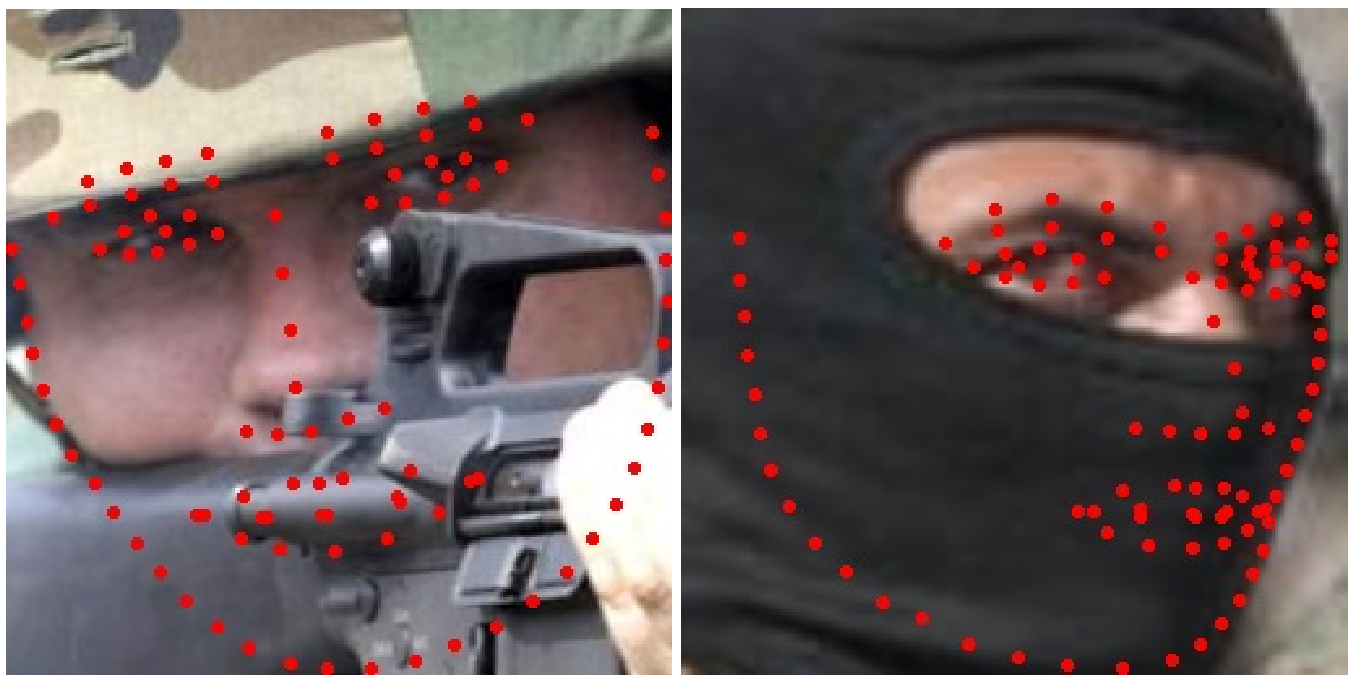}
    }         
    \subfigure[Rotation\label{fig:robust_rot}]{
    \includegraphics[width=0.23\linewidth]{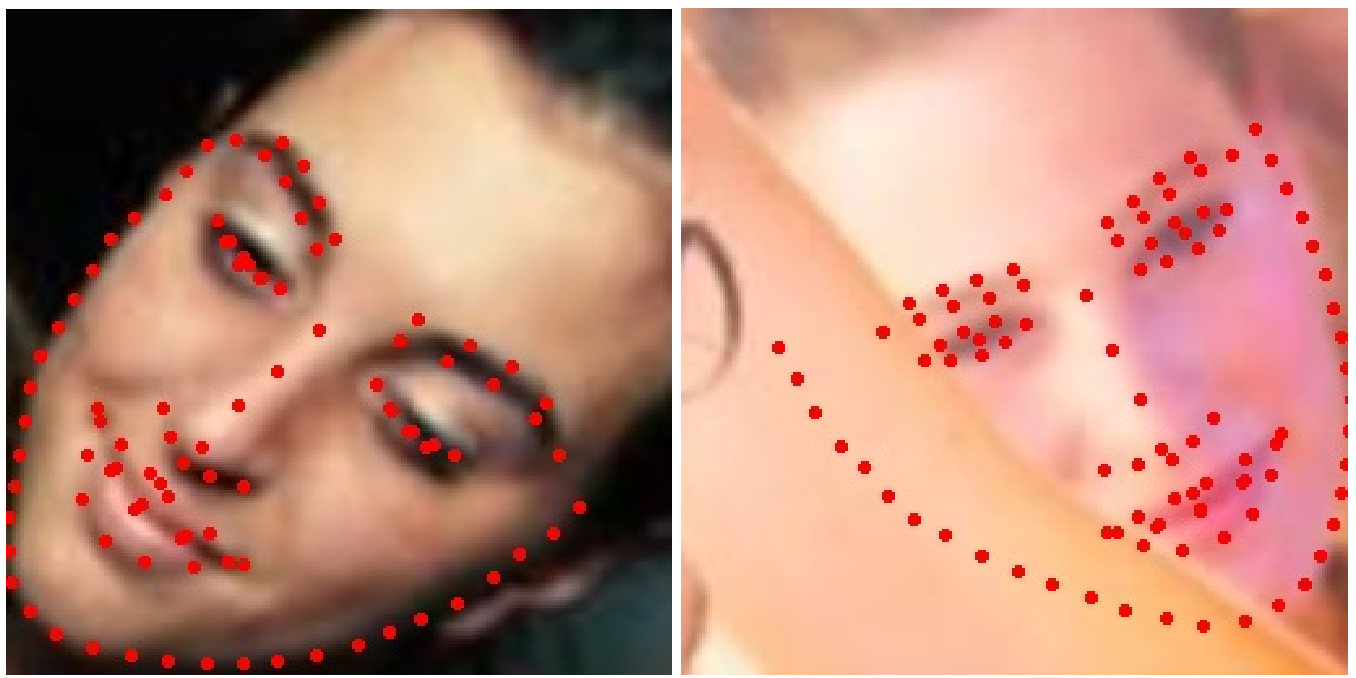}
    }      
    \subfigure[Blur\label{fig:robust_blur}]{
    \includegraphics[width=0.23\linewidth]{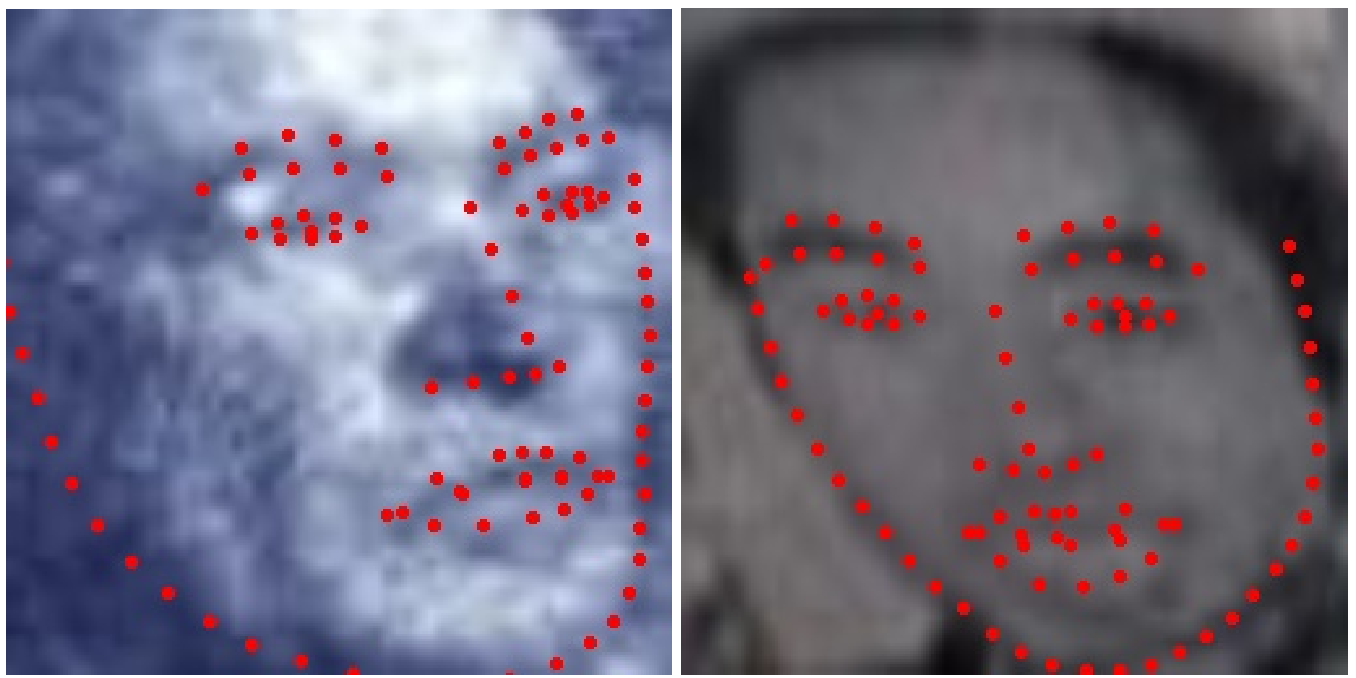}
    }
    \subfigure[Large pose\label{fig:robust_pose}]{
    \includegraphics[width=0.23\linewidth]{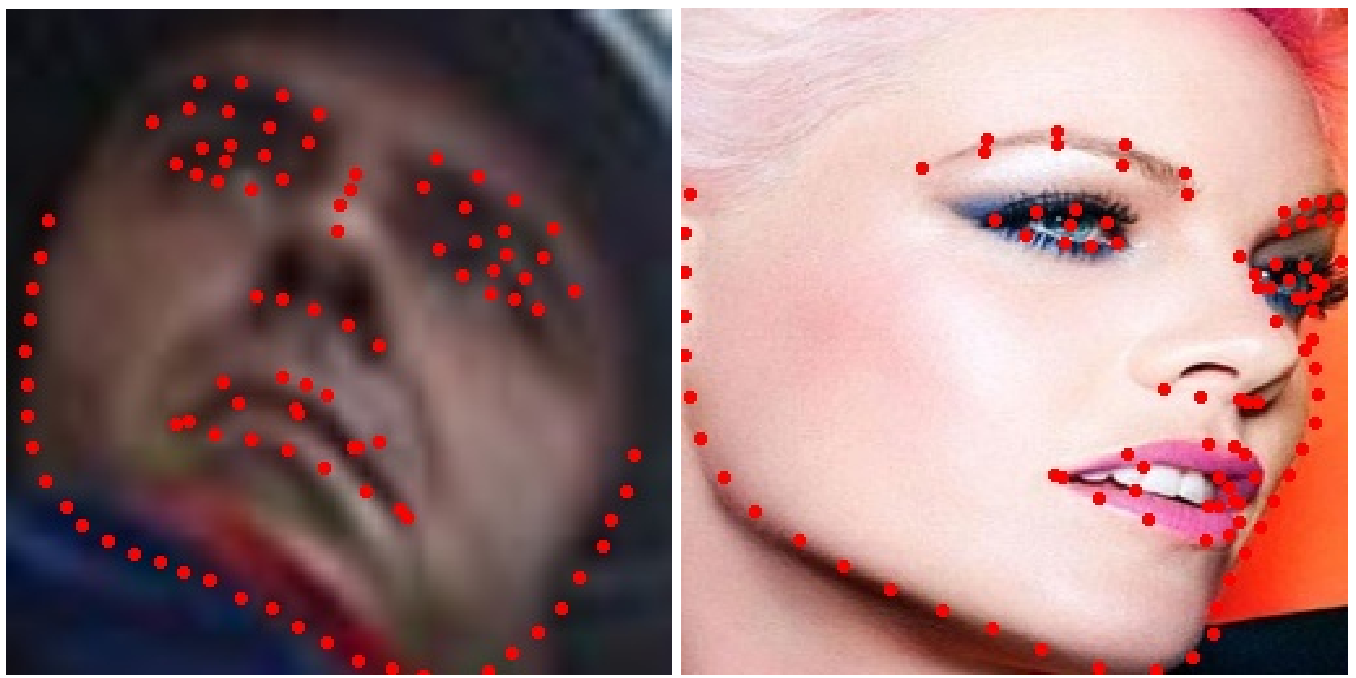}
    }  
    \vspace{-3mm}
    \caption{Sample images of WFLW with predictions from BarrelNet-50 under four typical scenarios. \label{fig:robust}}
\end{figure}

\begin{table}
\begin{minipage}{.45\linewidth}
    \centering
    \caption{NME (\%) of baseline ResNet-50 on WFLW with different number of encoder and decoder layers.}
    \label{tab:enc_num}
    \medskip
  \begin{tabular}{ccccccc}
    \toprule
    \multirow{2}{*}{Encoder Num.}  &  \multicolumn{4}{c}{Decoder Num.}                   \\
         \cmidrule(r){2-5}
        & 1     & 3  & 5     & 7  \\
    \midrule
    0  & 4.31 & 4.27  & 4.24  & 4.24       \\
    1  & 4.35 & 4.27  & 4.27  & 4.24       \\
    3  & - & 4.28  & 4.27  & 4.26       \\
    5  & - & -  & 4.28  & 4.26       \\
    7  & - & -  & -  & 4.29       \\
    \bottomrule
  \end{tabular}
\end{minipage}\hfill
\begin{minipage}{.45\linewidth}
    \centering
    \caption{NME (\%) on COFW-68. \textcolor{red}{Red} indicates best, and \textcolor{blue}{blue} for second best.}
    \label{tab:cofw_68}
    \medskip
\begin{tabular}{ccc}
    \toprule
    Model   &Backbone     & NME  \\
    \midrule
    LAB~\citep{WQY18}  & ResNet-18 & 4.62       \\
    ODN~\citep{ZSZ19}  & ResNet-18 & 5.30     \\
    AVS w/ SAN~\citep{QSW19}  & ITN-CPM & 4.43      \\
    PIPNet~\citep{JLS20}  & ResNet-18 & 4.55      \\
    DAG~\citep{LLZ20}  & HRNet-W18 & \textcolor{red}{\textbf{4.22}}      \\ 
    \midrule
    BarrelNet (ours)  & ResNet-18 & 4.41     \\
    BarrelNet (ours)  & ResNet-101 & \textcolor{blue}{\textbf{4.27}}     \\
    \bottomrule
  \end{tabular}
\end{minipage}
\end{table}

\subsection{Results}
\label{sec:4.2}

We compare our models with state-of-the-art methods on three benchmarks, and the results are summrized in Table~\ref{tab:results}. From the table, we can see that BarrelNet consistently outperforms our baseline model on all the datasets with different sizes of backbone, which indicates the effectiveness of DQInit and QAMem. On 300W full set, BarrelNet-101 achieves $3.09$, a very competitive result to the best existing ones (i.e., $3.07$ from AWing and $3.04$ from DAG). It is worth noting that AWing and DAG use stacked hourglass netwrok and HRNet as backbones respectively, both of which produce high-resolution maps for predictions, while BarrelNet only uses low-resolution feature maps which is more computationally efficient. On 300W challenging set, BarrelNet-101 obtains the second best result (4.60), which indicates its generalization capability on difficult and uncommon samples. Similarly, BarrelNet obtains the second best result (3.10) on COFW, indicating its robustness on heavy occlusions. In terms of WFLW, BarrelNet-101 achieves 4.20 NME, which is the new state of the art. The superior performance on WFLW is a solid proof for a practical model since the data distribution of WFLW is the closest to that of in-the-wild images among the three benchmarks. 

\subsection{Model analysis}
\label{sec:4.3}

To better understand the proposed model, we do more detailed analysis through extensive experiments.

\paragraph{More encoder and decoder layers.}
To observe how encoder and decoder influence our model, we run the baseline model with ResNet-50 on WFLW, varying the number of encoders and decoders. Table~\ref{tab:enc_num} summarizes the results. As can be seen, stacking more decoder layers improves the results while more encoders harms the results. This observation is not consistent with \citet{CMS20}, and we attribute it to the intrinsic difference of the two tasks. Specifically, we speculate that landmark detection primarily pays attention to the localization of single points while object detection also cares about the content encircled by the bounding boxes. Therefore, the clustering capability of encoders helps object detectors localize the whole objects more accurately while it may introduce spatial noises that hurt the localization of points.

\begin{figure}
\centering
    \subfigure[\label{fig:module_dec}]{
    \includegraphics[width=0.45\linewidth]{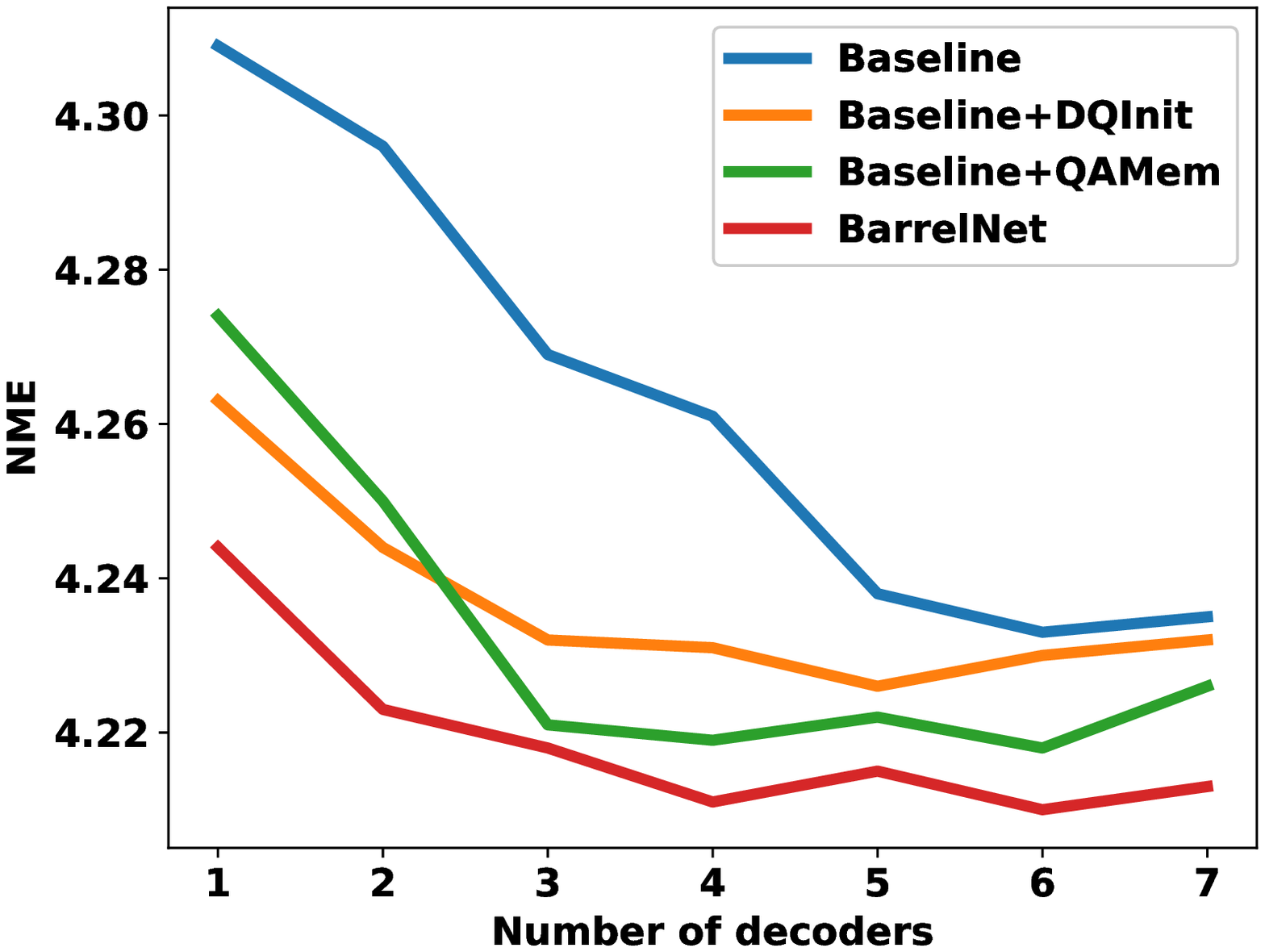}
    }         
    \subfigure[\label{fig:module_res}]{
    \includegraphics[width=0.45\linewidth]{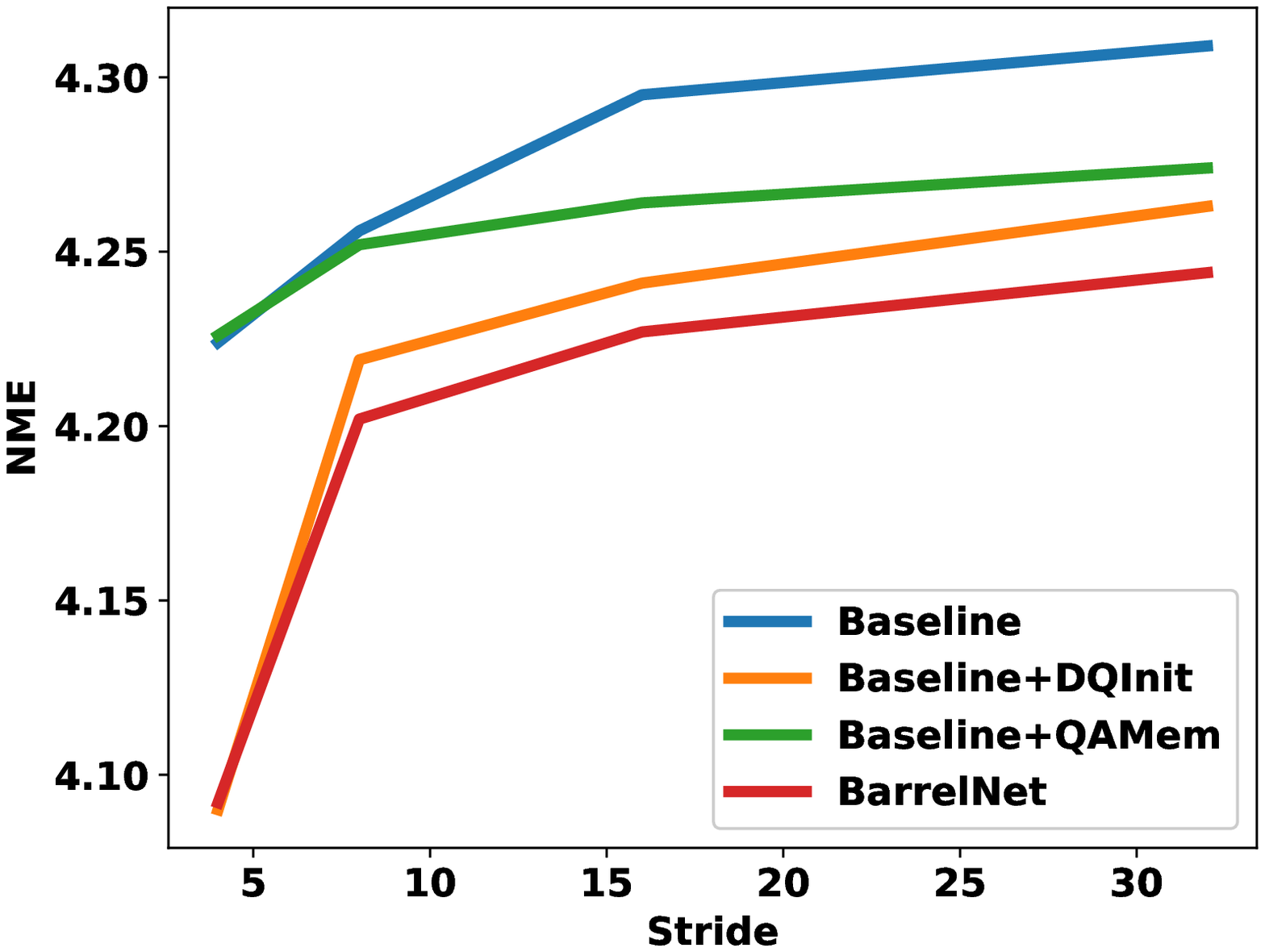}
    }        
    \vspace{-3mm}
    \caption{Comparisons between models w/ and w/o the propoosed modules on WFLW, using ResNet-50. (a) Varying number of decoders. (b) Varying resolution of feature maps. \label{fig:module}}
\end{figure}

\paragraph{Higher feature-map resolutions.}
According to prior works, high-resolution feature maps are beneficial to the performance of landmark detectors. Therefore, we run relevant experiments to see how resolutions affect our model. The default stride of ResNet-50 is 32. To obtain stride 16 and 8, we modify the first convolutional layer of stage 4/5 to be stride 1 and dilation 2, following \citep{CMS20}. Additionally, we adopt HRNet-W18, which is of stride 4. Figure~\ref{fig:module_res} shows the results on WFLW with different strides. As we can see, the accuracy of the baseline model (blue line) increases as the resolution becomes higher (i.e., stride becomes smaller). Therefore, higher-resolution feature maps do help improve the performance.

\begin{table}
  \caption{Comparisons between the baseline and BarrelNet on inference speed under various scenarios.}
  \label{tab:module_speed}
  \centering
  \begin{tabular}{lcccccc}
    \toprule
    Model   &Backbone     & Stride  & Decoder Num.     & NME  & FPS (GPU)  & FPS (CPU) \\
    \midrule
    Baseline  & ResNet-18 & 32  & 5  & 4.42  & 72.8  & 22.3     \\
    BarrelNet  & ResNet-18 & 32  & 1  & 4.42  & 137.2 (+\textbf{88.5}\%)  & 28.7 (+28.7\%)    \\
    \midrule
    Baseline  & ResNet-50 & 32  & 6  & 4.23  & 52.2  & 11.2     \\
    BarrelNet  & ResNet-50 & 32  & 1  & 4.24  & 90.5 (+\textbf{73.4}\%) & 14 (+25.0\%)     \\
    \midrule
    Baseline  & ResNet-50 & 8  & 1  & 4.26  & 85  & 4.4     \\
    BarrelNet  & ResNet-50 & 32  & 1  & 4.24  & 90.5 (+6.5\%)  & 14 (+\textbf{218.2}\%)    \\
    \midrule
    Baseline  & HRNet-W18 & 4  & 1  & 4.22  & 21.3  & 4.2     \\
    BarrelNet  & ResNet-101 & 32  & 1  & 4.20  & 55.3 (+\textbf{159.6}\%)  & 8.8 (+\textbf{109.5}\%)    \\
    \bottomrule
  \end{tabular}
\end{table}

\paragraph{Ablation study.}

To see how each module contributes to the improvement against the baseline, we do an ablation study with ResNet-50 on WFLW. From the results in Figure~\ref{fig:module_dec}, when the number of decoder is 1, we see that DQInit and QAMem reduce the NME from 4.31 to 4.26 and 4.27 respectively. When the two modules are used together, the NME is further reduced to 4.24. Therefore, the two modules both are effective for improving model performance.

\paragraph{Module analysis.}

In this part, we analyze the two proposed modules to understand their characteristics. First of all, we observe the effectiveness of DQInit and QAMem separately when the number of decoder layers and feature-map resolutions increase. As can be seen in Figure~\ref{fig:module_dec}, the NME of baseline decreases as the number of decoder increases, saturated around 6. When DQInit is added, there are considerable reductions on NME, but the reductions become negligible when decoder number is larger than 5. For QAMem, the NME reductions are consistent with different numbers of decoders, but the reductions after 5 decoders are also smaller. As expected, when the two modules are used together, it yields the best results across different numbers of decoders. Figure~\ref{fig:module_res} shows the results with different feature resolutions. From the figure, we observe that DQInit can improve the results consistently while QAMem is only effective for large strides such as 32 and 16. Again, using the two modules together achieves the best results. Based on the above experiments, we find that DQInit acts as extra decoder layers while QAMem brings more features related to resolutions.

Secondly, we examine the computational improvement of the proposed model under various scenarios. From the first two rows in Table~\ref{tab:module_speed}, we can see that BarrelNet-18 with one decoder achieves the same NME as the baseline with five decoders. As a consequence, BarrelNet-18 is 88.5\% and 28.7\% faster on GPU and CPU respectively. Similarly, BarrelNet-50 is 73.4\% and 25.0\% faster on GPU and CPU respectively. On the other hand, compared to the baseline with stride 8, BarrelNet-50 achieves slightly better result (4.24 vs. 4.26), while being more than $3\times$ faster on CPU. When HRNet with stride 4 is used for the baseline, the result of BarrelNet-101 is again slightly better (4.20 vs. 4.22), and its speed is $2.6\times$ and $2.1\times$ faster on GPU and CPU respectively. It is worth saying that the computation of more decoders is relatively slow on GPU while higher resolutions is slow on CPU, which could be a guideline for practical deployment. For example, if a model has been decided to deploy on GPU, then we can reduce the stride to obtain better accuracy with a little extra cost, if needed. On the contrary, more decoder layers could be added for a better accuracy if it is on lightweight devices such as CPU. 

\begin{figure}
\centering
    \subfigure[GPU\label{fig:speed_gpu}]{
    \includegraphics[width=0.48\linewidth]{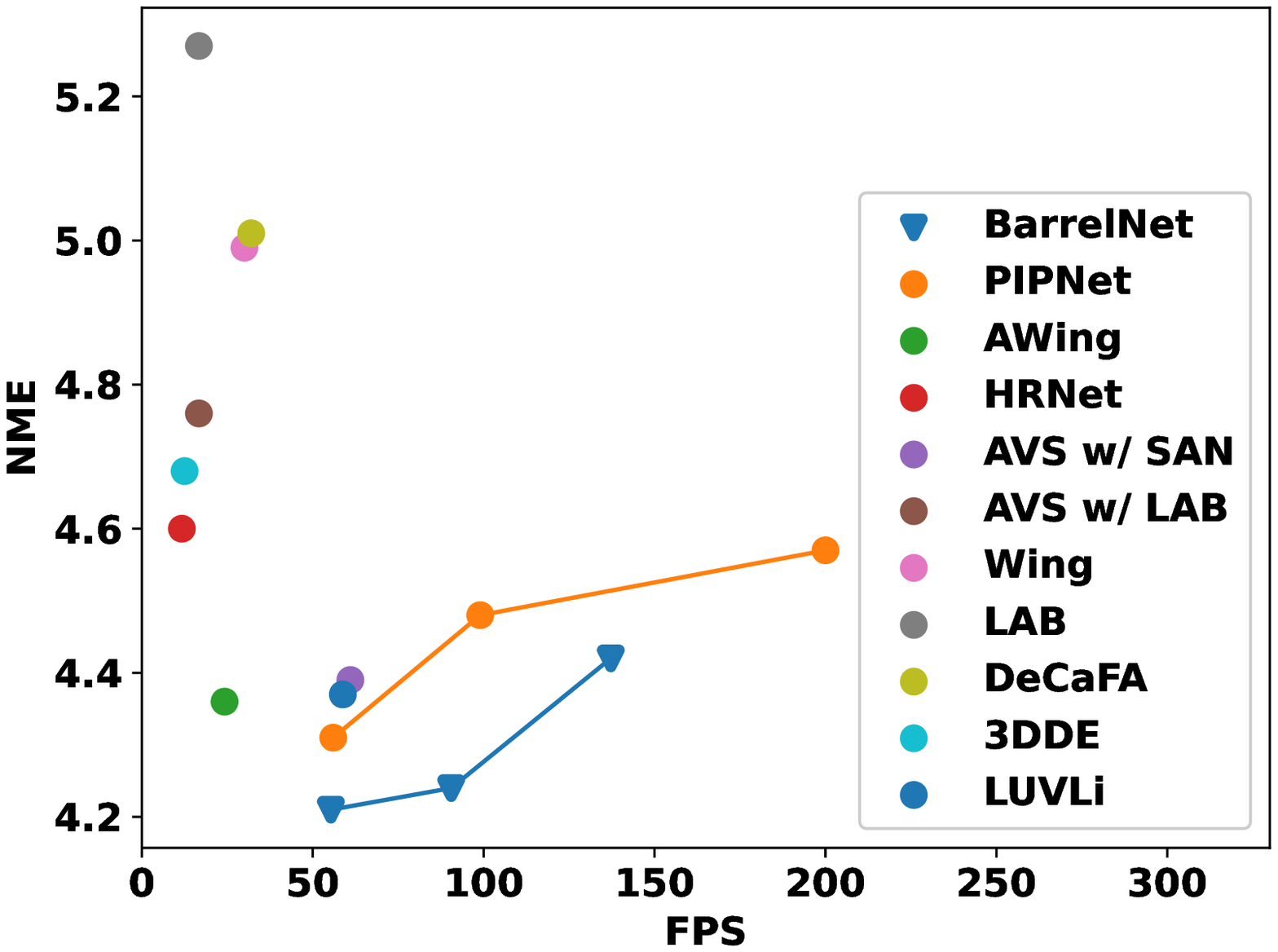}
    }         
    \subfigure[CPU\label{fig:speed_cpu}]{
    \includegraphics[width=0.48\linewidth]{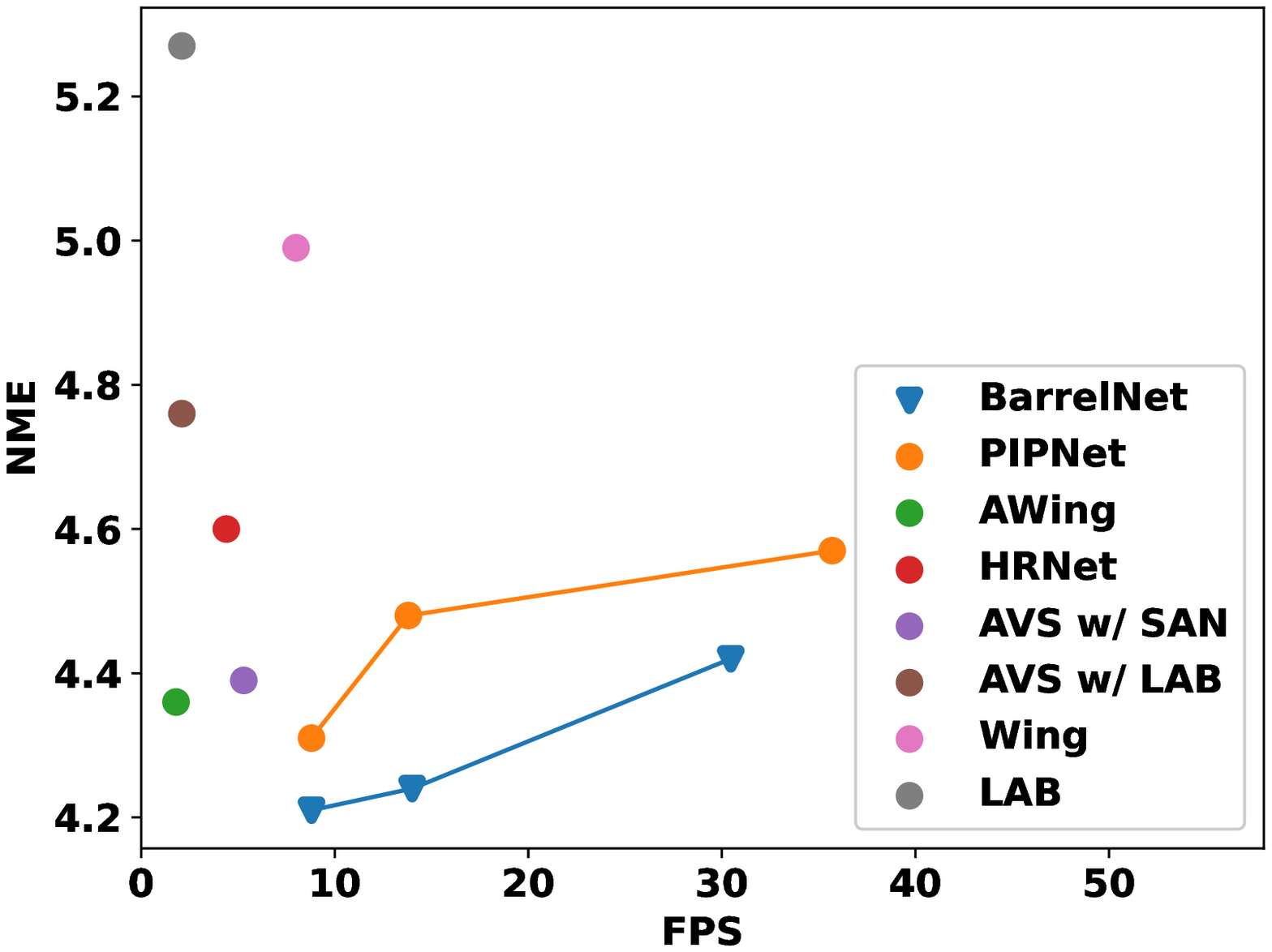}
    }        
    \vspace{-3mm}
    \caption{Comparison with the existing methods in terms of speed-accuracy trade-off on WFLW. \label{fig:speed}}
\end{figure}

\paragraph{Generalization.}

To quantitatively evaluate the generalization capability of our model, we report the results on COFW-68 following previous works~\citep{WQY18,ZSZ19,QSW19,JLS20,LLZ20}. Table~\ref{tab:cofw_68} shows the results on COFW-68, where the models are all trained on 300W training set. As we can see, BarrelNet obtains the best result (4.41) among the methods with ResNet-18. With ResNet-101, BarrelNet achieves 4.27, which is the second best result on COFW-68. We believe the competitive performance under the cross-data evaluation setting is a convincing indication of the generalization capability of our model.

\paragraph{Robustness.}
Next, we show the structural robustness of our model qualitatively. Figure~\ref{fig:robust} shows some sample images of WFLW under various scenarios with predicted landmarks from BarrelNet-50. From the figure, we see that the model still outputs meaningful geometries when there are heavy occlusions, rotations, blur, or large poses, thanks to the self-attention mechanism among queries. 

\paragraph{Efficiency.}
Finally, we demonstrate the advantage of our model on efficiency. Following \citep{JLS20}, we compare the speed-accuracy trade-off on WFLW using GPU and CPU respectively. A model being closer to the bottom right corner represents a better speed-accuracy trade-off. As can be seen in Figure~\ref{fig:speed_gpu} and \ref{fig:speed_cpu}, BarrelNet obtains the best trade-off on both GPU and CPU. Notably, BarrelNet-101 maintains the best performance on WFLW, while still running at 50+ FPS on GPU.

With both quantitative and qualitative results, we have demonstrated the practical advantages of our model on accuracy, generalization, structural robustness, and efficiency. Besides, we believe our model is relatively easy to train as it is end-to-end trainable and requires minimum hyperparameters.

\section{Discussion}
\label{sec:5}

Despite being practical on several aspects, there are a couple of issues from BarrelNet to be addressed. Firstly, like other transformer-based models~\citep{VSP17,CMS20}, BarrelNet requires longer training time than CNN-based models. To mitigate it, a sparse attention module could be used as in \citep{ZSL21} to speed up training convergence. Another problem of BarrelNet is the size of parameters it introduces. DQInit and QAMem introduce 4.4M and 9M parameters respectively, which considerably outperform the size of one decoder layer (1.6M). Although speed is a more essential factor than model size in practice, it would be better to further compress the model size to make it more lightweight. We see the above issues as a furture work.  

\section{Conclusion}
\label{sec:6}

In this work, we propose BarrelNet, a practical model for facial landmark detection. The model is designed to be accurate, efficient, structurally robust, generalizable, and end-to-end trainable by introducing a transformer decoder layer and two novel modules. The decoder is used as the detection head of the landmark detector with task-specific adaptations. DQInit is proposed to dynamically initialize the queries of decoder from the inputs, which boosts the accuracy with negligible extra cost. Moreover, we propose QAMem to enhance the discriminative ability of queries on low-resolution feature maps by utilizing separate memories. The competitive performance on three benchamrks indicates the superiority of BarrelNet in accuracy. Additionally, the model analysis with extensive experiments demonstrates the effectiveness and characteristics of the novel modules as well as our model's advantages in multiple practical aspects.

%\section*{References}

{\small
\bibliographystyle{abbrvnat.bst}
\bibliography{mybib}
}

\end{document}